\documentclass[10pt,twocolumn,letterpaper]{article}

\usepackage{cvpr}              
\definecolor{cvprblue}{rgb}{0.21,0.49,0.74}
\usepackage[pagebackref,breaklinks,colorlinks,allcolors=cvprblue]{hyperref}

\usepackage{algorithm}
\usepackage{algorithmic}
\usepackage{capt-of}

\usepackage{graphicx}
\usepackage{amsmath}
\usepackage{amsfonts}
\usepackage{multirow}
\usepackage{colortbl}
\usepackage{pifont}
\usepackage{booktabs}
\newcommand{\cmark}{\ding{51}}%
\newcommand{\xmark}{\ding{53}}%

\usepackage{xcolor}
\definecolor{my1}{RGB}{174,32,18} %
\definecolor{my2}{RGB}{188,62,3}  %
\definecolor{my3}{RGB}{204,102,2} %
\definecolor{my4}{RGB}{238,181,47}  %
\definecolor{my5}{RGB}{255,228,175} %

\def\paperID{} 
\def\confName{CVPR}
\def\confYear{2026}

\begin{document}

\title{TR2M: \underline{T}ransferring Monocular \underline{R}elative Depth \underline{to} \underline{M}etric Depth with Language Descriptions and Dual-Level Scale-Oriented Contrast}


\author{
Beilei Cui\textsuperscript{1}\footnotemark[1]\quad
Yiming Huang\textsuperscript{1}\footnotemark[1]\quad
Long Bai\textsuperscript{1}\quad
Hongliang Ren\textsuperscript{1}\footnotemark[2] \\
\textsuperscript{1}The Chinese University of Hong Kong, Hong Kong\\
{\tt\small \{beileicui,yhuangdl,b.long\}@link.cuhk.edu.hk, hlren@ee.cuhk.edu.hk} \\
\footnotesize
\textsuperscript{*}Equal contribution.\quad
\textsuperscript{\dag}Corresponding author. \\
}

\twocolumn[{
\maketitle

\begin{center}
  \includegraphics[width=0.9\textwidth]{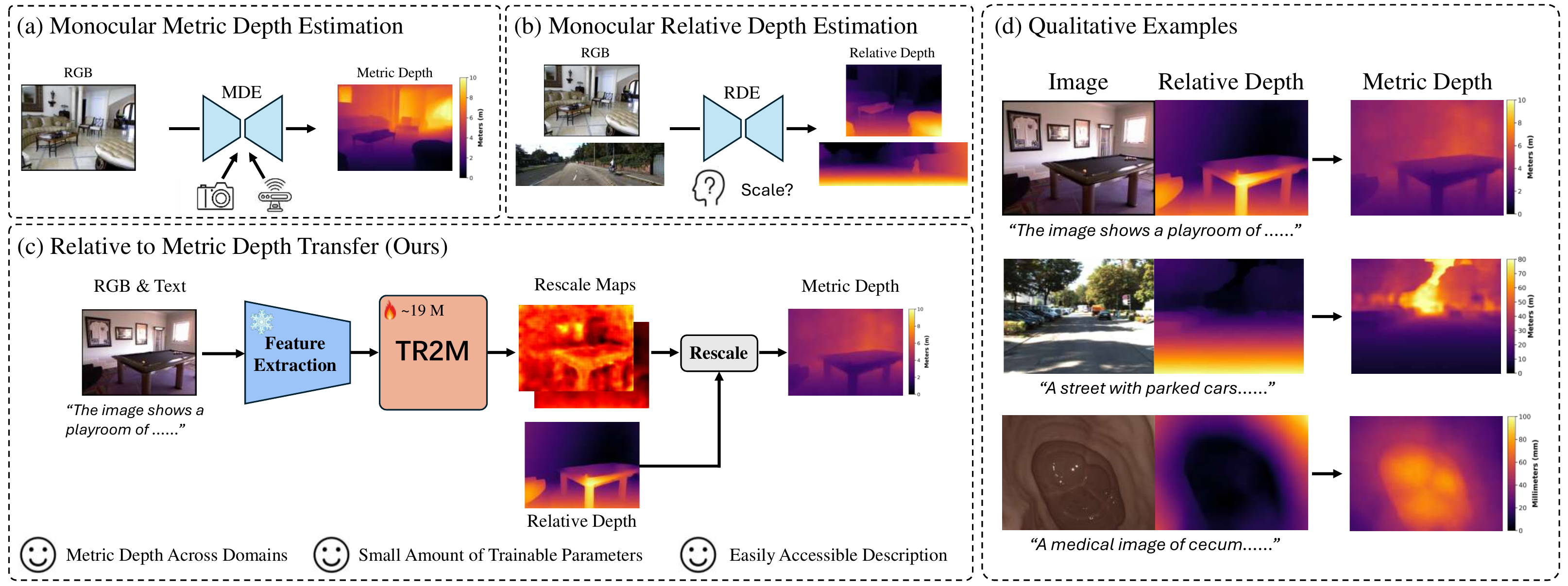}

  \captionof{figure}{Illustration of motivation. (a) Metric Depth Estimation typically restricts to a single domain or relies on camera parameters or sensors to enhance generalization. (b) Relative Depth Estimation generalizes better but is ambiguous in scale. (c) We therefore seek to transfer generalizable relative depth to metric depth by pixel-wise rescaling maps given image and readily accessible text description, which succeeds in obtaining metric depth for various domains with one lightweight trainable architecture. (d) Qualitative examples.}
  \label{fig:first}
\end{center}


}]

\begin{abstract}
This work presents a generalizable framework to transfer relative depth to metric depth. Current monocular depth estimation methods are mainly divided into metric depth estimation (MMDE) and relative depth estimation (MRDE). MMDEs estimate depth in metric scale but are often limited to a specific domain. MRDEs generalize well across different domains, but with uncertain scales that hinder downstream applications. To this end, we aim to build up a framework to solve scale uncertainty and transfer relative depth to metric depth. Previous methods used language as input and estimated two factors for conducting rescaling. Our approach, TR2M, utilizes both text descriptions and images as inputs and estimates two rescale maps to transfer relative depth to metric depth at the pixel level. Features from two modalities are fused with a cross-modality attention module to better capture scale information. A strategy is designed to construct and filter confident pseudo metric depth for more comprehensive supervision. We also develop dual-level scale-oriented contrastive learning to utilize depth distribution as guidance to enforce the model learning about intrinsic cues consistent with the scale distribution. TR2M only exploits a small number of trainable parameters to train on datasets in various domains and experiments not only demonstrate TR2M’s great performance in seen datasets but also reveal superior zero-shot capabilities on five unseen datasets. We show the huge potential in pixel-wise transferring relative depth to metric depth with language assistance instead of large-size metric depth models with large amounts of training data. Code is available at: \url{https://github.com/BeileiCui/TR2M}.
\end{abstract}    
\section{Introduction}
\label{sec:intro}

Monocular depth estimation (MDE) plays a crucial role in comprehending the 3D layout of environments depicted in 2D images and finds numerous applications in robotics~\cite{dong2022towards, wofk2019fastdepth}, autonomous driving~\cite{peng2020ida, chen2015deepdriving}, endoscopic surgery~\cite{huang2024endo, liu2019dense} and virtual reality technologies~\cite{el2019survey, du2020depthlab}. Predicting a dense depth map from a single image is ill-posed without additional sensors because any points along the ray for a pixel with different depth values can yield the same image coordinate.

Recent research on MDE focuses mainly on two aspects: monocular metric depth estimation (MMDE) and monocular relative depth estimation (MRDE) as shown in Figure~\ref{fig:first} (a) and (b). MMDE aims at predicting a dense depth map in a specific scale (often in meters). A limitless range of outputs leads early works' performance to commonly degrade much on test data with diverse scales~\cite{zhao2024metric}. Recent advancements have been made to rectify such a situation by leveraging additional prior conditions such as camera information~\cite{guizilini2023towards, yin2023metric3d} or scene information~\cite{bhat2023zoedepth}. Alternatively, MRDE estimates the depth map in a unified scale. The affine-invariant loss enables optimization on vast amounts of data, resulting in excellent generalization across domains. However, the scale uncertainty of MRDE also limits its practical applicability where a determined scale is needed, such as in robot planning and navigation.

We consider addressing scale uncertainty for transferring relative depth to metric depth efficiently to construct an MMDE pipeline with as good generalization capabilities across different domains as MRDE methods. Language descriptions effectively predict scene scale for relative-to-metric depth conversion (MMDE), leveraging object-scene correlations (e.g., vehicles/houses outdoors imply larger scale)~\cite{zengrsa}. However, two major problems have affected the performance of previous language description based methods in predicting scene scales: 1) Large regions of erroneous prediction still occur in state-of-the-art relative depth models, therefore a single factor for transforming relative depth map to metric scaled depth map will keep or even magnify such inaccuracy; 2) Same objects with different distributions and proportions result in inconsistent overall depth, but the language descriptions may be the same or very similar. Meanwhile, scale consistency at the feature level is often overlooked. Pixels with similar depth and scale may have low similarity in pixel-level features, leading to instability of overall depth estimation.

In light of the above observations, we develop a framework to predict pixel-wise transformation parameters to \textbf{T}ransfer a monocular \textbf{R}elative depth map \textbf{to} a \textbf{M}etric depth map based on the image and corresponding text description termed \textbf{TR2M} as shown in Figure~\ref{fig:first} (c). The image itself serves as input for the depth model, while the language description is readily available, eliminating the need for additional sensors or calibrations. The image and text feature embeddings are acquired with separate frozen encoders. A lightweight network with cross-modality attention takes two feature embeddings as inputs and predicts two maps functioning as a scale map and a shift map. Then, the metric depth can be obtained by a simple pixel-wise linear transformation based on a scale and shift map, which is supervised by the ground truth metric depth. To encourage scale continuity and pixel-wise consistency in regions without ground-truth supervision, we also construct pseudo metric depth by aligning relative and ground truth metric depth and use a threshold to select high-confidence pseudo depths as the supervision signal. We further innovatively propose a Dual-Level Scale-Oriented Contrast to enhance the feature consistency with the same depth scale. Image-wise features are enforced to be aligned with their corresponding scales and pixel-wise features are classified by their depth distribution and maximize the distance among features with different distributions and minimize the distance among features with the same distributions. We demonstrate that TR2M yields results similar to aligning relative depth through a linear transformation with the ground truth, and it surpasses some SOTA metric depth estimation methods with fewer trainable parameters, offering improved generalization across diverse domains.

We summarize our main contributions as follows:

\begin{itemize}
    \item[1)] We propose a novel framework that leverages image and text descriptions to predict rescale maps to transfer monocular relative depth map to metric depth map across multiple domains.
    
    \item[2)] We introduce a Dual-Level Scale-Oriented Contrast Learning strategy by classifying depth based on its distribution to enforce global scale consistency.
    
    \item[3)] Extensive experiments have demonstrated the superior performance of our proposed method on multiple scenes. To the best of our knowledge, we are the first to employ text descriptions to perform pixel-wise transformation from relative to metric depth.

\end{itemize}
\section{Related Work}
\label{sec:related_work}
\subsection{Monocular Depth Estimation}

Recent Monocular Depth Estimation (MDE) comprises metric (MMDE) and relative (MRDE) depth estimation. MMDE models learn to predict pixel-wise depth maps in metric scale supervised by ground truth metric depth~\cite{bhat2021adabins, bhat2022localbins, fu2018deep, wong2020targeted}. Remarkable advances have been made in network architecture~\cite{piccinelli2023idisc, yuan2022neural, laina2016deeper, yang2021transformer}, geometry modeling~\cite{Han_2023_ICCV, Zhao_2023_ICCV, Shao_2023_ICCV, Yang_2023_ICCV} and image priors~\cite{Rodriguez-Puigvert_2023_ICCV, patni2024ecodepth}. However, these methods typically restrict their setting to specific domains, resulting in performance degradation under domain shifts and limited generalization to unobserved environments. Recent studies have investigated MMDE models with improved generalization ability with the help of camera intrinsic-based modeling~\cite{antequera2020mapillary, yin2023metric3d, piccinelli2024unidepth, piccinelli2025unik3d} or direct integration of camera intrinsic~\cite{guizilini2023towards, facil2019cam}. These methods often need fine-tuning to adapt to specific domains or rely on particular sensors or cameras as prior knowledge. In contrast, by training on large-scale datasets with multifarious scales and domains accompanied by scale-invariant loss function~\cite{Ranftl2021, Ranftl2022, yang2024depth, ke2024repurposing}, relative depth models have been proven to have better generalization across different scenes. As a trade-off, the metric scale can not be recovered by relative depth models, which limits downstream applications to some extent. TR2M aims to break through such a predicament by aligning highly generalizable relative depth into a metric scale efficiently to construct a universal metric depth estimation pipeline.

\subsection{Language Involved Depth Estimation}
Vision-Language Models (VLM)~\cite{caron2021emerging, oquab2023dinov2, radford2021learning} develop a holistic grasp of both textual and visual content by undergoing pre-training on varied datasets, thereby establishing a robust foundation that enhances performance in subsequent downstream tasks~\cite{yang2024binding, you2024calibrating, zhang2022pointclip, zhu2023pointclip}. Recently, some researchers have attempted to leverage the power of VLM for monocular depth estimation~\cite{zhang2025hybrid, cai2025depthlm}. DepthCLIP~\cite{zhang2022can} utilizes CLIP's~\cite{radford2021learning} semantic features combined with a depth projection mechanism to enable zero-shot adaptation, translating semantic language understanding into monocular depth estimation tasks. Wordepth~\cite{zeng2024wordepth} utilizes the Variational Auto-Encoder (VAE) to learn the geometry distribution from the text description. Further advancements were made based on DepthCLIP by learnable prompts~\cite{hu2024learning} or tokens~\cite{auty2023learning}. ScaleDepth~\cite{zhu2024scale} designed bin and scale queries to estimate the relative depth and scale factor for metric depth. However, the designed correlations between text and depth are indirect and too scratchy to represent precise depth information.

\subsection{Relative to Metric Depth Transfer}
With the advancement in relative depth estimation models in recent years, researchers have started to study the transfer of powerful RDE models to metric depth scale. Efforts have been made to utilize left-right stereo consistency~\cite{wu2022toward}, camera geometric embedding~\cite{guizilini2023towards}, or test-time adaptation~\cite{zhao2024metric} to capture scale information. Most state-of-the-art methods train a relative depth model on diverse datasets first and then fine-tune the encoder with a metric head to specific metric datasets~\cite{ranftl2021vision, Ranftl2022, yang2024depth, yang2025depth}. RSA~\cite{zengrsa} estimates scale factor and shift factor from text input to globally rescale relative depth to metric depth for different scenes. However, metric heads are limited to specific datasets, and complex environments may mislead text descriptions to inaccurate scale estimation. More direct and integrated relative to metric depth transformation methods are still under exploration. 

\subsection{Contrastive Learning for Depth Estimation}
Contrastive Learning has achieved substantial progress in learning feature representations for unsupervised learning tasks~\cite{he2020momentum, chen2020improved, wu2018unsupervised}. Pixel-level contrastive learning emerges for dense prediction with or without labels as guidance~\cite {zhao2021contrastive, wang2021exploring, jin2022exploring, wang2021dense}. WCL utilizes windows to form positive and negative pairs from depth feature maps for contrastive learning~\cite{fan2023contrastive}. D4RD~\cite{li2024region} developed a robust depth estimation method by enforcing consistency among sample noises with different perturbations. Li et al.~\cite{li2024region} model depth features with a Gaussian distribution to conduct region-wise contrastive learning. Current contrastive learning methods for depth estimation are mainly unsupervised learning or supervised by other types of labels, without utilizing depth itself as a supervisory signal.
\section{Method}
\label{sec:method}
\subsection{Overview}

Figure~\ref{fig:mainframework} shows an overview framework of the proposed TR2M. Taking an RGB Image $I \in \mathbb{R}^{H \times W \times 3}$ and its text description $L$ as input, TR2M aims to estimate two maps: scale map $A \in \mathbb{R}^{H \times W}$ and shift map $B \in \mathbb{R}^{H \times W}$ to rescale its corresponding relative depth $D_{r} \in \mathbb{R}^{H \times W}$ to metric depth $\hat{D}_{m} \in \mathbb{R}^{H \times W}$. Note that $D_{r} \in \mathbb{R}^{H \times W}$ is generated by a pre-trained relative depth model. Besides optimizing $\hat{D}_{m}$ with the ground truth depth map $D^{gt}_{m}$, we also perform linear regression to find the scale and shift factor $\tilde{\alpha}$ and $\tilde{\beta}$ that minimizes pseudo rescaled metric depth $D^{pseudo}_{m}$ with $D^{gt}_{m}$. If the quality of $D^{pseudo}_{m}$ is good enough based on an evaluation threshold, it is also used as a supervision for $\hat{D}_{m}$. Furthermore, we also propose dual-level scale-oriented contrast learning to enforce feature consistency guided by depth distribution, thus improving the scale perception ability.

\subsection{Scale and Shift Map Estimation}
\label{sec:sub_rescale}

\begin{figure*}[t]
\centering
\includegraphics[width=0.9\linewidth]{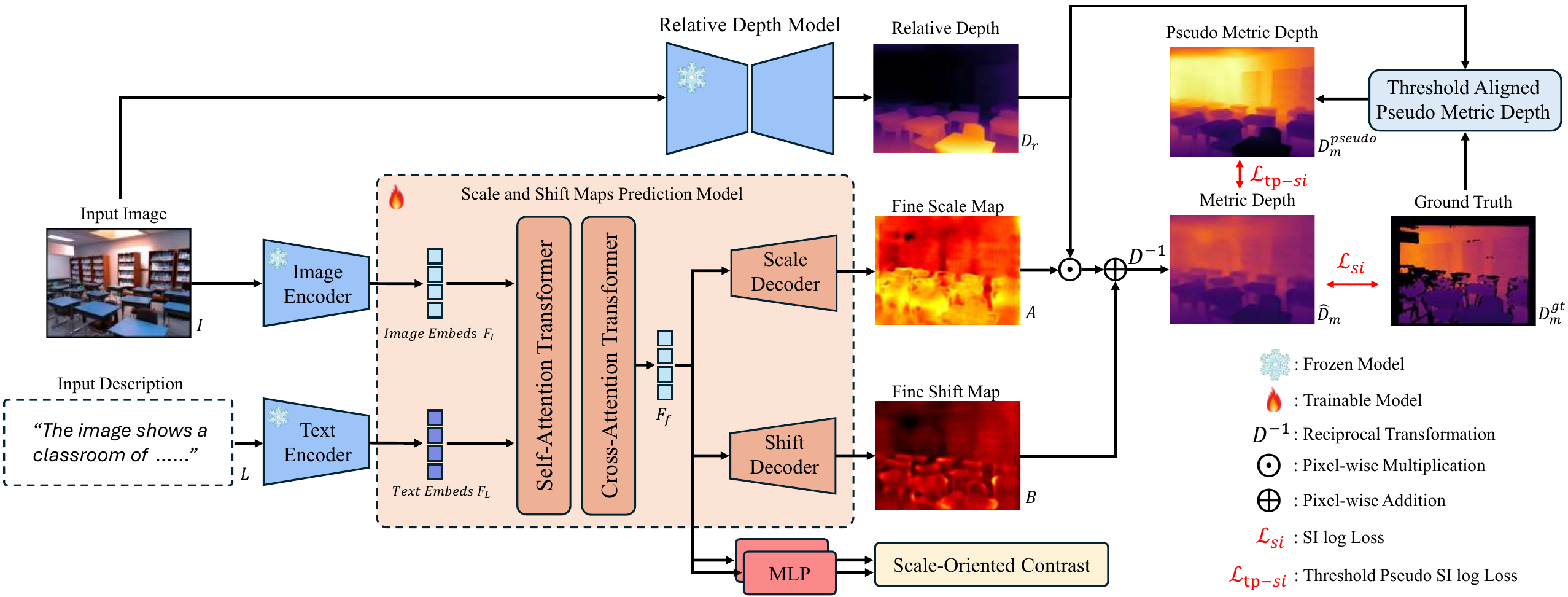}
\caption{Illustration of the proposed TR2M framework. Image and text embedding features are first obtained with separate frozen encoders, and a cross-modality attention module is proposed to integrate them. The rescale maps are predicted with different decoders to transfer relative depth to metric depth. Ground truth depth map and aligned pseudo metric depth map are utilized. Scale-Oriented Contrast is applied to the final embedding features.}
\label{fig:mainframework}
\end{figure*}

\paragraph{Network Architecture.} Both images and textual descriptions can provide clues about scale, but a single modality alone with loss of overall layout or average scale remains ambiguous. We endeavor to comprehensively utilize both image and text to learn more accurate pixel-level scale information as shown in Figure~\ref{fig:mainframework}. To be specific, with the RGB image $I \in \mathbb{R}^{H \times W \times 3}$ and text description $L$ as inputs, we first extract image feature $F_{I} \in \mathbb{R}^{HW \times D}$ and text feature $F_{L} \in \mathbb{R}^{1 \times D}$ separately with a frozen image encoder and a frozen text encoder. 

To capture information effectively from both feature embeddings, we introduce a cross-modality attention module as in Eq. (\ref{cross-attn1}). To maintain the pixel-level output, we make the image feature embedding query to access information from the key and value of the text feature embedding. Image feature embedding first applies self-attention and aggregates text information with skip connection, as in Eq. (\ref{cross-attn2}), to obtain the final feature embedding $F_{out} \in \mathbb{R}^{HW \times D}$.
\begin{equation}
\text {Attn}_{cm}^i\left(Q_I, K_i, V_i\right)=\operatorname{softmax}\left(\frac{Q_I K_i^T}{\sqrt{d}}\right) \cdot V_i, i \in\{I, L\}, \label{cross-attn1}
\end{equation}
\begin{equation}
F_{out}=F_{I}+\text{Attn}_{cm}^I+\text{Attn}_{cm}^L . 
\label{cross-attn2} 
\end{equation}
Then, two lightweight decoder heads as in DPT~\cite{Ranftl2021} are applied to generate the fine scale map $A \in \mathbb{R}^{H \times W}$ and fine shift map $B \in \mathbb{R}^{H \times W}$ with $A = \operatorname{Scalehead} \left (  F_{f}\right ), B = \operatorname{Shifthead} \left (  F_{f}\right )$ to make pixel-wise adjustments to the scale and shift. We aim at transferring relative depth to metric depth, so we assume a pre-trained frozen RDE model exists to generate the relative depth map $D_{r} \in \mathbb{R}^{H \times W}$. The final metric depth can be obtained by rescaling $D_{r}$ as below:
\begin{equation}
\hat{D}_{m} = \frac{1}{A \odot D_{r} + B} \text{,}
\end{equation}
where $\odot$ denotes element-wise multiplication. A Scale-Invariant log loss~\cite{eigen2014depth} is utilized as follows:
\begin{equation}
\mathcal{L}_{\mathrm{si}}\left(\hat{D}_{m}, D^{gt}_{m}\right)=\frac{1}{H W} \sum_{i=1}^{HW} \epsilon_i^2-\frac{\lambda}{(H W)^2}\left(\sum_{i=1}^{HW} \epsilon_i\right)^2,
\end{equation}
where $\epsilon_i = log\hat{D}_{m}(i) - logD^{gt}_{m}(i)$ and $i$ denotes the index of pixels.

\paragraph{Pseudo Metric Depth with Threshold.} The sparsity of ground truth metric depth results in partial pixels having no supervision to learn scale information. To conduct more comprehensive scale supervision, we leverage pseudo metric depth by aligning relative and metric depth on a least-squares criterion. To be specific, let $\tilde{\alpha}, \tilde{\beta} \in \mathbb{R}^{1}$ denote single estimators for scale and shift, respectively. Following~\cite{Ranftl2022}, a meaningful alignment could be made based on a least-squares criterion as below:
\begin{equation}
(\tilde{\alpha}, \tilde{\beta})  =\arg \min _{\tilde{\alpha}, \tilde{\beta}} \sum_{i=1}^{HW}\left(\tilde{\alpha} D_{r}(i)+\tilde{\beta}-D^{gt}_{m}(i)\right)^2, 
\label{leastsquares} 
\end{equation}
The aligned pseudo metric depth can be determined by: $D^{pseudo}_{m}  =\tilde{\alpha} D_{r}+\tilde{\beta}$ with the above $\tilde{\alpha}$ and $\tilde{\beta}$, which can be efficiently solved in closed form by rewriting Eq. (\ref{leastsquares}) as a least-squares problem (details are shown in the Supplementary). We then use threshold accuracy $\delta_1$ (percentage of $\max \left(D^{gt}_{m} / D^{pseudo}_{m}, D^{pseudo}_{m} / D^{gt}_{m}\right)<1.25$) metric as a condition. If $\delta_1$ of $D^{pseudo}_{m}$ is greater than a preset threshold value, we consider $D^{pseudo}_{m}$ as a credible pseudo depth and use it for supervision, which can be written as:
\begin{equation}
\mathcal{L}_{\mathrm{tp-si}}\left(\hat{D}_{m}, D^{pseudo}_{m}\right) = \mathbf{1}(\delta _{1} > \rho ) \cdot \mathcal{L}_{\mathrm{si}}\left(\hat{D}_{m}, D^{pseudo}_{m}\right),
\end{equation}
where $\mathbf{1}(\cdot)$ is the indicator function, $\delta _{1}$ is the threshold accuracy of $D^{pseudo}_{m}$ and $\rho$ is the preset threshold value.

\subsection{Dual-Level Scale-Oriented Contrast}

\begin{figure}[t]
\centering
\includegraphics[width=0.9\linewidth]{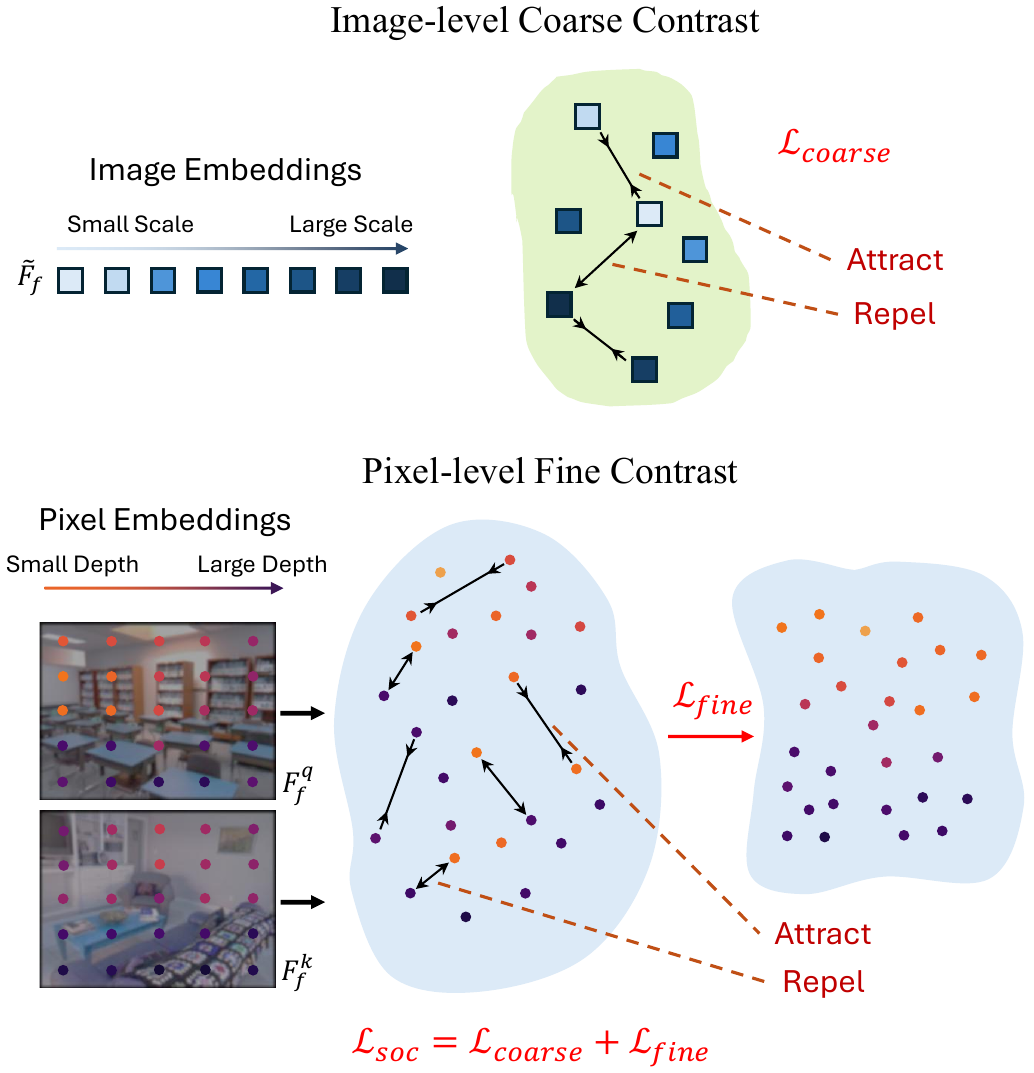}
\caption{Illustration of the proposed Dual-Level Scale Oriented Contrast Learning. The scale-oriented contrast enables embedding features more consistent with the scale and depth distribution within different levels, thus enhancing scale perception capability.}
\label{fig:main_2}
\end{figure}

\label{sec:sub_contrast}
As demonstrated by Depth Anything~\cite{yang2024depth}, semantic encoders such as DINOv2~\cite{oquab2023dinov2} often produce similar features for distinct object regions (e.g., car front vs. back). In contrast, depth estimation requires distinguishing potentially large variations, even between adjacent pixels. To enhance cross-domain generalization, models must capture the intrinsic relationships governing depth values. As shown in Figure~\ref{fig:main_2}, this approach enforces embedding similarity for features with similar scale and depth distributions and dissimilarity for those with divergent distributions. 

\textbf{Image-Level Coarse Contrast.}
We first conduct an image-level coarse contrast to ensure that the overall image embeddings are aligned with their scale distribution. For each training iteration, we first select an embedding set $\textbf{F}$ composed with image embeddings $\tilde{\textbf{F}} = \left [ \tilde{F}^{0}_{f}, \tilde{F}^{1}_{f}, ... , \tilde{F}^{M-1}_{f} \right ]$ where $\tilde{F}_{f}$ represents the fine feature embedding for an image obtained by average pooling on $F_{f}$. For simplicity, we assume that the pseudo scale factors $\tilde{\alpha}_{i}$ for each feature embeddings in the set are ordered ($\forall i, j \quad with \quad i< j,\quad \alpha_{i}< \alpha_{j}$). We then introduce a contrastive loss to ensure that the embeddings with the same scale levels are closer than embeddings with different scale levels:

\begin{equation}
\begin{gathered}
\mathcal{L}_{coarse}=-\frac{1}{M} \sum_{i=0}^{M-1}\log \frac{\exp \left(S_i^{c-p}\right)}{\exp \left(S_i^{c-p}\right)+\exp \left(S_i^{c-n}\right)}, \\
S_i^{c-p}=\frac{1}{\left|\textbf{F}\right|} \sum_{F_1, F_2\in \textbf{F}}\sum_{|i - j| < t}\left\langle F_1^{i} \cdot F_2^{j}\right\rangle, \\
S_i^{c-n}=\frac{1}{\left|\textbf{F}\right|(M - 1)} \sum_{F_1, F_2\in \textbf{F}}\sum_{|i - j| \geq t}\left\langle F_1^{i} \cdot F_2^{j}\right\rangle,
\end{gathered}
\end{equation}

where $\left< \cdot \right>$ is the cosine similarity function, $t$ is the scale similarity factor that controls and distinguishes similar scales, $S^ {c-p}$ and $S^{c-n}$ denote positive pairs and negative pairs.  It ensures that the
learned features reflect the natural scale ordering for the scenes of the images.

\textbf{Pixel-Level Fine Contrast.}
In addition to image-level coarse contrast, we leverage depth ground truth for pixel-level fine contrast to align pixel embeddings with the depth distribution. Given an image $I^{q} \in \mathbb{R}^{H \times W \times 3}$ as a query sample and a set of key samples, we train the model by distinguishing the positive ones from the negative ones at the pixel level. Following~\cite{jin2022exploring, he2020momentum}, we decouple the model into two symmetric branches that are updated by back-propagation and exponential moving average (EMA)~\cite{hunter1986exponentially} separately. Each branch contains a proposed rescale model and a projector to map the feature $F_{f}$ into a lower dimension. Query and key samples are fed into different branch to predict query and key feature maps $F^{q}_{f}, F^{k}_{f} \in \mathbb{R}^{h \times w \times D_{c}}$ and pixel-level feature embedding can be obtained with $F^{q}_{f}(i)$ and $F^{k}_{f}(j)$ for the $i$-th and $j$-th pixel, respectively.

We convert the depth map into a classification map based on the depth distribution. To be specific, consider $\mathcal{C}$ classes of distribution, the classification map can be calculated by:
\begin{equation}
Y = \text{round}(\frac{d_{max} - D}{d_{max} - d_{min}} \times |\mathcal{C}|),
\end{equation}
where $D$ is the depth map, $d_{min}$ and $d_{max}$ are the minimum and maximum depth value. The corresponding class map for query and key $Y^{q}, Y^{k} \in \mathbb{R}^{h \times w \times |\mathcal{C}|}$ can then be obtained by resolution reduction and one-hot encoding. $\left\{ F^{q}_{f}(i), Y^{q}_{f}(i)\right\}$ represents query pixel $i$ and $\left\{ F^{k}_{f}(j), Y^{k}_{f}(j)\right\}$ represents key pixel $j$, separately. For query pixel $i$, we then compute a binary selection mask: $\mathcal{M} \subseteq\{0,1\}^{\frac{H}{S} \times \frac{W}{S}}$ for each key feature map $F^{k}$ which is determined by the class: $\mathcal{M}_j=\mathbf{1}\left[Y^{q}(i)=Y^{k}(j)\right]$, where $\mathbf{1}(\cdot)$ is indicator function. $F^{k}(j)$ is grouped into positive samples when two pixels have the same class, denoted as $F^{k}_{f}(j)^{+}$, and grouped into negative sample otherwise, denoted as $F^{k}_{f}(j)^{-}$. Our scale-oriented contrast is then formulated to maximize similarities for positive pairs and minimize them for negative pairs:
\begin{equation}
\begin{gathered}
\mathcal{L}_{fine}=-\frac{1}{hw} \sum_{i=1}^{hw}\log \frac{\exp \left(S_i^{f-p}\right)}{\exp \left(S_i^{f-p}\right)+\exp \left(S_i^{f-n}\right)}, \\
S_i^{f-p}=\frac{1}{\left|\mathcal{P}_i\right|} \sum_{j+\in \mathcal{P}_i}\left\langle F^{q}_{f}(i) \cdot F^{k}_{f}(j)^{+}\right\rangle, \\
S_i^{f-n}=\sum \frac{1}{\left|\mathcal{N}^n_i\right|} \sum_{j-\in \mathcal{N}^n_i}\left\langle F^{q}_{f}(i) \cdot F^{k}_{f}(j)^{-}\right\rangle,
\end{gathered}
\end{equation}
where $\mathcal{P}_i$ and $\mathcal{N}_i$ represent the sets of pixel embeddings corresponding to positive and negative samples for pixel i, respectively. The features are reorganized based on the depth distribution, enhancing the model's generalization ability for scale perception.

The final scale-oriented loss consists of the two level losses $\mathcal{L}_{soc} = \mathcal{L}_{coarse} + \mathcal{L}_{fine}$.

\begin{table*}[t]
\caption{Quantitative results on NYUv2. I, L, R, and M refer to Image, Language, Relative, and Metric, denoting the input and output type of the method. "Domain Adap" denotes Domain Adaptation. "FT" means fine-tuned. "Median" denotes using median scaling for each image. "Linear Fit" refers to using ground truth to optimize scale and shift for each image. "Global" refers to optimizing a single scale and shift for the entire dataset. \colorbox{gray!25}{Gray} results denote alignment with ground truth is used. $^*$ denotes reproduced in our environment. We highlighted the \colorbox{my1!70}{first}, \colorbox{my2!50}{second}, \colorbox{my3!40}{third}, \colorbox{my4!40}{fourth} and \colorbox{my5!40}{fifth} best results.}
  \centering
  \resizebox{0.9\textwidth}{!}{
  \begin{tabular}{l|c|c|c|c|ccc|ccc}
    \toprule
    Model & Type & Scaling & Train Images & Train Params & $\delta_1 \uparrow$ & $\delta_2 \uparrow$ & $\delta_3 \uparrow$ & $Abs Rel\downarrow$ &  $log_{10}\downarrow$ &  $RMSE\downarrow$  \\
    \midrule
    \multicolumn{11}{l}{\textit{Direct Metric Depth Estimation Methods}} \\
    DA$^*$~\cite{yang2024depth} & I $\rightarrow$ M & Domain Adap & - & 25M & 0.957 & 0.995 & 0.999 & 0.075 & 0.033 & 0.273 \\
    DA V2~\cite{yang2025depth} & I $\rightarrow$ M & Domain Adap & - & 25M & 0.969 & 0.996 & 0.999 & 0.073 & 0.032 & 0.261 \\ 
    ZoeDepth~\cite{bhat2023zoedepth} & I $\rightarrow$ M & Domain Adap & - & 345M & 0.951 & 0.994 & 0.999 & 0.077 & 0.033 & 0.282 \\
    UniK3D$^*$~\cite{piccinelli2025unik3d} & I $\rightarrow$ M & Zero-shot & 8M & 359M & 0.958 & 0.991 & 0.996 & 0.072 & 0.031 & 0.268 \\
    Metric3Dv2$^*$~\cite{hu2024metric3d} & I $\rightarrow$ M & Zero-shot & 16M & 1011M & 0.980 & 0.997 & 0.999 & 0.067 & 0.030 & 0.260 \\
    UniDepth$^*$~\cite{piccinelli2024unidepth} & I $\rightarrow$ M & Zero-shot & 3M & 347M & 0.981 & 0.997 & 1.000 & 0.072 & 0.030 & 0.229 \\
    \midrule
    \multicolumn{11}{l}{\textit{Indirect Metric Depth Estimation Methods}} \\
    \rowcolor{gray!25}
    DA~\cite{yang2024depth} & R $\rightarrow$ M & Median & 64M & 25M & 0.480 & 0.734 & 0.886 & 0.353 & 0.135 & 1.743 \\
    \rowcolor{gray!25}
    DA~\cite{yang2024depth} & R $\rightarrow$ M & Linear Fit & 64M & 25M & 0.969 & 0.995 & 0.999 & 0.055 & 0.024 & 0.260 \\
    DA~\cite{yang2024depth} & R $\rightarrow$ M & Global & 64M & 25M & 0.630 & 0.926 & 0.987 & 0.199 & 0.087 & 0.646 \\
    DepthCLIP~\cite{zhang2022can} & I,L $\rightarrow$ M & Zero-shot & - & - & 0.394 & 0.683 & 0.851 & 0.388 & 0.156 & 1.167 \\
    DepthLM~\cite{cai2025depthlm} & I,L $\rightarrow$ M & Zero-shot & - & - & \colorbox{my4!40}{0.824} & \colorbox{my5!40}{0.933} & \colorbox{my5!40}{0.990} & \colorbox{my4!40}{0.134} & \colorbox{my4!40}{0.068} & \colorbox{my4!40}{0.401} \\
    ScaleDepth$^*$~\cite{zhu2024scale} & I,L $\rightarrow$ M & Domain Adap & - & 109M & \colorbox{my3!40}{0.913} & \colorbox{my3!40}{0.989} & \colorbox{my2!50}{0.998} & \colorbox{my3!40}{0.099} & \colorbox{my3!40}{0.041} & \colorbox{my2!50}{0.329} \\     
    WorDepth$^*$~\cite{zeng2024wordepth} & I,L $\rightarrow$ M & Domain Adap & - & 137M & \colorbox{my2!50}{0.926} & \colorbox{my2!50}{0.990} & \colorbox{my2!50}{0.998} & \colorbox{my2!50}{0.090} & \colorbox{my2!50}{0.040} & \colorbox{my3!40}{0.330} \\
    RSA$^*$~\cite{zengrsa} & (I,L) R $\rightarrow$ M & Rescale factors & 102K & 4.7M & \colorbox{my5!40}{0.752} & \colorbox{my4!40}{0.964} & \colorbox{my5!40}{0.992} & \colorbox{my5!40}{0.156} & \colorbox{my5!40}{0.071} & \colorbox{my5!40}{0.528} \\
    TR2M (Ours) & (I,L) R $\rightarrow$ M & Rescale maps & 102K & 19M & \colorbox{my1!70}{0.954} & \colorbox{my1!70}{0.996} & \colorbox{my1!70}{0.999} & \colorbox{my1!70}{0.082} & \colorbox{my1!70}{0.035} & \colorbox{my1!70}{0.293} \\
    \bottomrule
  \end{tabular}}
  \label{tab:nyu}
\end{table*}

\begin{table*}[th]
  \caption{Quantitative results on KITTI. Details about notations can be found at Table~\ref{tab:nyu}.}
  \centering
  \resizebox{0.9\textwidth}{!}{
  \begin{tabular}{l|c|c|c|c|ccc|ccc}
    \toprule
    Model & Type & Scaling & Train Images & Train Params & $\delta_1 \uparrow$ & $\delta_2 \uparrow$ & $\delta_3 \uparrow$ & $Abs Rel\downarrow$ &  $RMSE_{log}\downarrow$ &  $RMSE\downarrow$ \\
    \midrule
    \multicolumn{11}{l}{\textit{Direct Metric Depth Estimation Methods}} \\
    DA$^*$~\cite{yang2024depth} & I $\rightarrow$ M & Domain Adap & - & 25M & 0.971 & 0.996 & 0.999 & 0.054 & 0.083 & 2.290 \\
    DA V2~\cite{yang2025depth} & I $\rightarrow$ M & Domain Adap & - & 25M & 0.973 & 0.997 & 0.999 & 0.053 & 0.081 & 2.235\\
    ZoeDepth~\cite{bhat2023zoedepth} & I $\rightarrow$ M & Domain Adap & - & 345M & 0.971 & 0.996 & 0.999 & 0.054 & 0.082 & 2.281\\
    UniK3D$^*$~\cite{piccinelli2025unik3d} & I$\rightarrow$ M & Zero-shot & 8M & 359M & 0.945 & 0.994 & 0.998 & 0.117 & 0.129 & 2.995 \\ 
    Metric3Dv2$^*$~\cite{hu2024metric3d} & I$\rightarrow$ M & Zero-shot & 16M & 1011M & 0.977 & 0.996 & 0.999 & 0.051 & 0.080 & 2.403 \\ 
    UniDepth$^*$~\cite{piccinelli2024unidepth} & I$\rightarrow$ M & Zero-shot & 3M & 347M & 0.985 & 0.998 & 0.999 & 0.043 & 0.067 & 1.820 \\ 
    \midrule
    \multicolumn{11}{l}{\textit{Indirect Metric Depth Estimation Methods}} \\
    \rowcolor{gray!25}
    DA~\cite{yang2024depth} & R $\rightarrow$ M & Median  &  64M & 25M & 0.925 & 0.986 & 0.996 & 0.091 & 0.129 & 3.648 \\
    \rowcolor{gray!25}
    DA~\cite{yang2024depth} & R $\rightarrow$ M & Linear Fit & 64M &  25M & 0.944 & 0.991 & 0.998 & 0.077 & 0.122 & 3.190 \\
    DA~\cite{yang2024depth} & R $\rightarrow$ M & Global & 64M &  25M & 0.663 & 0.932 & 0.981 & 0.191 & 0.228 & 5.273 \\
    DepthCLIP~\cite{zhang2022can} & I,L $\rightarrow$ M & Zero-shot & - & - & 0.465 & 0.713 & 0.867 & 0.343 & 0.305 & 7.583 \\
    DepthLM~\cite{cai2025depthlm} & I,L $\rightarrow$ M & Zero-shot & - & - & \colorbox{my5!40}{0.765} & \colorbox{my5!40}{0.912} & \colorbox{my5!40}{0.988} & \colorbox{my5!40}{0.156} & \colorbox{my5!40}{0.193} & \colorbox{my5!40}{4.356} \\
    ScaleDepth$^*$~\cite{zhu2024scale} & I,L $\rightarrow$ M & Domain Adap & - &  109M & \colorbox{my1!70}{0.968} & \colorbox{my1!70}{0.996} & \colorbox{my1!70}{0.999} & \colorbox{my1!70}{0.058} & \colorbox{my1!70}{0.086} & \colorbox{my1!70}{2.235} \\
    WorDepth$^*$~\cite{zeng2024wordepth} & I,L $\rightarrow$ M & Domain Adap & - &  137M & \colorbox{my2!50}{0.965} & \colorbox{my1!70}{0.996}	& \colorbox{my1!70}{0.999} & \colorbox{my2!50}{0.066} & \colorbox{my2!50}{0.088} & \colorbox{my3!40}{2.356} \\
    RSA$^*$~\cite{zengrsa} & (I,L) R $\rightarrow$ M & Rescale factors & 102K & 4.7M & \colorbox{my4!40}{0.786} & \colorbox{my4!40}{0.967} & \colorbox{my4!40}{0.995} & \colorbox{my4!40}{0.147} & \colorbox{my4!40}{0.179} & \colorbox{my4!40}{4.143} \\
    TR2M (Ours) & (I,L) R $\rightarrow$ M & Rescale maps & 102K & 19M & \colorbox{my2!50}{0.965} & \colorbox{my1!70}{0.996} & \colorbox{my1!70}{0.999} & \colorbox{my2!50}{0.066} & \colorbox{my3!40}{0.093} & \colorbox{my2!50}{2.328} \\
    \bottomrule
  \end{tabular}}
  
  \label{tab:kitti}
\end{table*}

\subsection{Training strategies}
\label{sec:sub_trainning}

Besides the above-mentioned loss function, we also implement edge-aware smoothness loss~\cite{godard2017unsupervised} $\mathcal{L}_{es}$ to enforce the smoothness property of the scale and shift maps. We train the framework in an end-to-end mode with a combination of all optimization objectives:
\begin{equation}
\mathcal{L} = \lambda_1\mathcal{L}_{si} + \lambda_2\mathcal{L}_{tp-si} + \lambda_3\mathcal{L}_{soc} + \lambda_4\mathcal{L}_{es}.
\end{equation}
More detailed settings are shown in the Supplementary.

\section{Experiments}
\label{sec:experiments}

\paragraph{Datasets}
Our primary datasets for training and evaluation for TR2M are NYUv2~\cite{silberman2012indoor}, KITTI~\cite{geiger2012we}, VOID~\cite{wong2020unsupervised} and C3VD~\cite{bobrow2023} ranging from outdoor and indoor scenes to surgical scenarios to enable depth transfer across various domains with one model. To demonstrate generalization ability, we evaluate zero-shot performance on five real-world and synthetic datasets: SUN RGB-D~\cite{song2015sun}, iBims-1~\cite{koch2018evaluation}, HyperSim~\cite{roberts2021hypersim}, DIODE Outdoors~\cite{vasiljevic2019diode}, and SimCol~\cite{rau2023bimodal}. We followed ~\cite{zengrsa} to generate text description with LLaVA v1.6 Vicuna and Mistral~\cite{liu2024improved}. More details about the datasets are presented in the Supplementary.

\paragraph{Evaluation Metrics}
Following~\cite{bhat2023zoedepth, zengrsa, zeng2024wordepth, yang2024depth, yang2025depth}, we evaluate our model in metric depth with commonly used metrics for depth estimation tasks, including $Abs Rel$, $RMSE$, $RMSE_{log}$, $log_{10}$, $\delta_1$, $\delta_2$ and $\delta_3$ where the first four are lower-is-better and the last three are higher-is-better.

\paragraph{Implementation Details}
All experiments were conducted on NVIDIA RTX4090 GPUs. We use the AdamW~\cite{loshchilov2017decoupled} optimizer with a batch size of 8. Learning rate is set to $1 \times 10^{-5}$ with a decay of $0.9$ every epoch. We train our model for 20 epochs and the hyperparameters $\lambda$, $\lambda_1$, $\lambda_2$, $\lambda_3$ and $\lambda_4$ are set to $0.15$, $1$, $0.5$, $0.1$ and $0.01$, respectively. The number of learnable Scale Embeddings is 256. We use Depth Anything-Small as our frozen relative depth model. We use the Vit-L/14 model from CLIP~\cite{radford2021learning} and the ViT-L model from DINOv2~\cite{oquab2023dinov2} as our frozen text and image encoder, respectively. Note that all evaluation metrics on our proposed method and baselines are conducted \textbf{without any post-processing} like scale alignment or median scaling for strict evaluation on the metric scale.

\begin{table*}[t]
  \caption{Zero-shot metric depth estimation. The first three test sets are indoor scenes; the fourth is for outdoor scenes and the last one is a surgical scene dataset. Note that ZoeDepth and DA Single use models are trained on NYUv2 for indoor evaluation, models trained on KITTI for outdoor evaluation, and models trained on C3VD for surgical evaluation. Other compared baselines and our model utilize one model trained on multiple datasets for all zero-shot evaluations. }
  \centering
  \resizebox{0.9\textwidth}{!}{
  \begin{tabular}{@{}l|c|c|cccccccccc|c@{}}
    \toprule
    \multirow{2}{*}{Method} & \multirow{2}{*}{Backbone} & \multirow{2}{*}{Train Images} & \multicolumn{2}{c}{SUN RGB-D} & \multicolumn{2}{c}{iBims-1} & \multicolumn{2}{c}{HyperSim} & \multicolumn{2}{c}{DIODE Outdoor} & \multicolumn{2}{c|}{SimCol} & \multirow{2}{*}{Rank} \\
    & & & $Abs Rel\downarrow$ & $\delta_1 \uparrow$ & $Abs Rel\downarrow$ & $\delta_1 \uparrow$ & $Abs Rel\downarrow$ & $\delta_1 \uparrow$ & $Abs Rel\downarrow$ & $\delta_1 \uparrow$ & $Abs Rel\downarrow$ & $\delta_1 \uparrow$ & \\
    \midrule
ZoeDepth &BeiT$_{384}$-Large & - & \colorbox{my5!40}{0.520} & \colorbox{my4!40}{0.545} & \colorbox{my4!40}{0.169} & \colorbox{my5!40}{0.656} & 0.407 & 0.302 & \colorbox{my4!40}{0.814} & 0.237 & \colorbox{my3!40}{0.372} & \colorbox{my3!40}{0.438} & \colorbox{my5!40}{4.70} \\ 
DA Single & ViT-Large & - & \colorbox{my3!40}{0.500} & \colorbox{my1!70}{0.660} & \colorbox{my1!70}{0.150} & \colorbox{my3!40}{0.714} & \colorbox{my5!40}{0.363} & \colorbox{my4!40}{0.361} & \colorbox{my3!40}{0.794} & \colorbox{my4!40}{0.288} & \colorbox{my2!50}{0.302} & \colorbox{my1!70}{0.553} & \colorbox{my2!50}{2.80} \\ 
DA Mix & ViT-Large & 102K & \colorbox{my4!40}{0.503} & \colorbox{my3!40}{0.582} & \colorbox{my3!40}{0.158} & \colorbox{my4!40}{0.705} & 0.412 & 0.296 & 0.842 & 0.181 & \colorbox{my4!40}{0.382} & \colorbox{my4!40}{0.432} & 5.00 \\ 
Unidepth-V-L & ViT-Large & 3M & 0.538 & 0.443 & 0.384 & 0.217 & \colorbox{my2!50}{0.294} & \colorbox{my2!50}{0.545} & \colorbox{my1!70}{0.504} & \colorbox{my1!70}{0.635} & - & - & \colorbox{my4!40}{4.00} \\ 
UniK3D-V-S & ViT-Small & 8M & 0.680 & 0.234 & 0.504 & 0.415 & \colorbox{my3!40}{0.349} & \colorbox{my3!40}{0.525} & 1.678 & \colorbox{my3!40}{0.414} & - & - & 6.00 \\ 
UniK3D-V-L & ViT-Large & 8M & 0.564 & 0.382 & \colorbox{my5!40}{0.186} & \colorbox{my2!50}{0.735} & \colorbox{my1!70}{0.232} & \colorbox{my1!70}{0.660} & \colorbox{my5!40}{0.833} & \colorbox{my2!50}{0.603} & - & - & \colorbox{my3!40}{3.75} \\ 
RSA & ViT-Small & 102K & \colorbox{my2!50}{0.457} & \colorbox{my5!40}{0.527} & 0.266 & 0.450 & 0.461 & 0.230 & 0.852 & 0.244 & \colorbox{my5!40}{0.466} & \colorbox{my5!40}{0.162} & 5.80 \\ 
TR2M (Ours) & ViT-Small & 102K & \colorbox{my1!70}{0.451} & \colorbox{my2!50}{0.591} & \colorbox{my2!50}{0.154} & \colorbox{my1!70}{0.736} & \colorbox{my4!40}{0.357} & \colorbox{my4!40}{0.361} & \colorbox{my2!50}{0.673} & \colorbox{my5!40}{0.274} & \colorbox{my1!70}{0.284} & \colorbox{my2!50}{0.445} & \colorbox{my1!70}{2.40} \\
    \bottomrule
  \end{tabular}}
  \label{tab:zero-shot}
\end{table*}

\subsection{Comparison Results on Depth Transfer}
We first evaluate the metric depth results on NYUv2 and KITTI which are common indoor and outdoor benchmarks. We reproduced some methods for fair comparison that are marked with $^*$ in the tables. In Table~\ref{tab:nyu}, our method outperforms all the other language/transfer based methods across all evaluation metrics and even obtained better $\delta_2$ and $\delta_3$ compared to the results of DepthAnything with linear fit, which requires ground truth to be rescaled for NYUv2. In Table~\ref{tab:kitti}, our model also obtains satisfactory results on KITTI with two best and three second-best evaluation metrics. Two important facts and observations further demonstrate the superiority of our method. \textbf{First,} our method only contains 19M trainable parameters, which is much smaller than most other methods. RSA has a smaller model size but results in much worse performance for its single-factor way of rescaling. \textbf{Second,} methods requiring domain adaptation are limited to training and evaluating on a single domain. Methods capable of zero-shot evaluation, like UniK3D and Metric3Dv2, require training on a large amount of data. By contrast, our method is trained end-to-end with multiple domains utilizing only 102K images for training and obtained comparable results. These two aspects are further demonstrated in the next section. 

\begin{figure*}[t]
\centering
\includegraphics[width=0.9\linewidth]{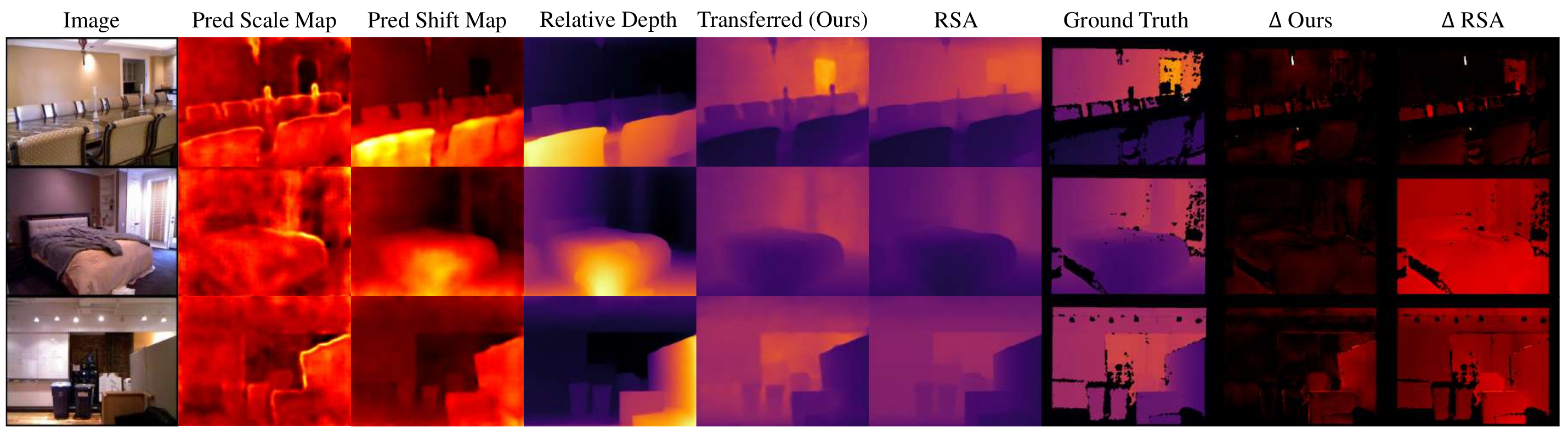}
\caption{Qualitative results on NYUv2. Our method consistently produces better predictions with much less error. $ \Delta $ denotes $Abs Rel$ ranging from lowest (Black) to highest (Red).}
\label{fig:vis_1}
\end{figure*}

\begin{figure}[t]
\centering
\includegraphics[width=0.9\linewidth]{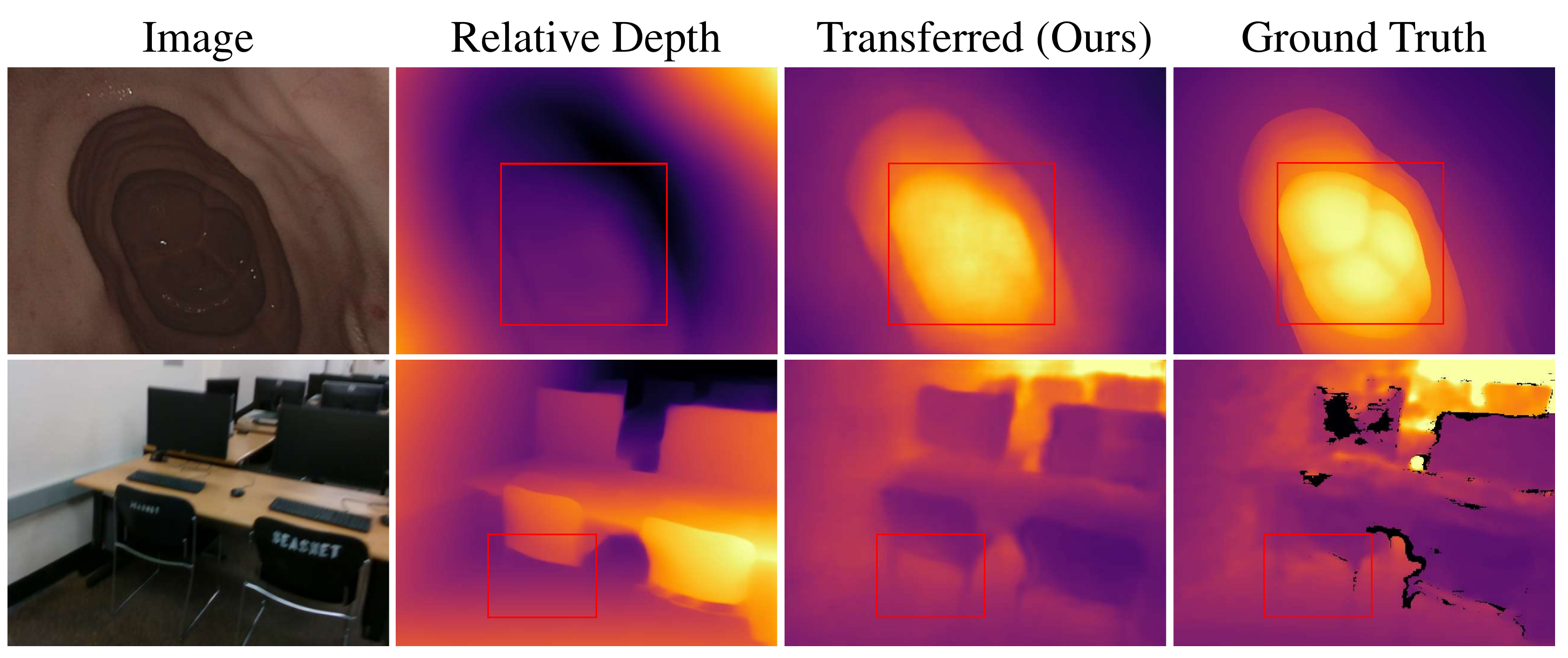}
\caption{The value of the relative depth map within the red rectangles is inconsistent with the ground truth. Our method can correct such errors when transferring to metric depth.}
\label{fig:vis_2}
\end{figure}

Figures~\ref{fig:vis_1} and \ref{fig:vis_2} show qualitative results: our method produces more accurate, consistently improved depth maps and, via pixel-level transformation, can correct erroneous regions in relative depth. Additional results and analysis are provided in the Supplementary.

\subsection{Zero-shot Generalization}
We also conduct zero-shot metric depth estimation on five unseen datasets. We fine-tuned DepthAnything with the four datasets we used for fair comparison. We adopt the same evaluation protocol for all baselines: predicted metric depths are evaluated directly against ground truth without any post-processing. As shown in Table~\ref{tab:zero-shot}, our TR2M results in the best average ranking compared to other methods. It is worth highlighting that our method achieves notable zero-shot performance with a ViT-Small ($\sim$ 25M parameters) backbone, whose scale is less than $1/10$ of most other compared baselines with large architectures ($\sim$ 345M parameters). Also, we only utilize 102K images for training, which is much smaller than other metric depth methods like UniK3D and UniDepth trained on a large amount of data. Training one DA model with mixed datasets (DA Mix in Table~\ref{tab:zero-shot}) degrades the performance, which is possibly because it relies on domain-specific decoders; mixing domains with large gaps may hinder its ability to perceive domain and scale. We aim to demonstrate that a metric depth model exhibiting strong generalization capabilities across domains need not rely on large-scale training datasets or separate, complex model architectures when provided with readily available image or text embeddings—features commonly accessible in multimodal tasks.

\subsection{Ablation Study}
Due to limited space, we only present the two most important ablation studies here. More ablation experiments and discussions are shown in the Supplementary.
\paragraph{Effectiveness of Proposed Modules.} We study the effect of our proposed modules, including the rescale map design, threshold pseudo metric depth supervision $\mathcal{L}_{tp-si}$, and scale-oriented contrast $\mathcal{L}_{soc}$. The results are summarized in Table~\ref{tab:ablation_module}.  \xmark \ in Rescale Map denotes using two single factors instead of maps to rescale, while \xmark \ for $\mathcal{L}_{tp-si}$ and $\mathcal{L}_{soc}$ simply refers to discarding the module. As expected, utilizing all the modules has the highest performance. Discarding $\mathcal{L}_{soc}$ results in worse performance, especially for zero-shot evaluation for the latter two datasets. Scale-oriented contrast enables the model to learn about knowledge of scale and depth distribution, therefore enhancing depth estimation accuracy. Discarding $\mathcal{L}_{tp-si}$ has limited effects on the seen dataset NYUv2 but degrades in the unseen datasets iBims-1 and DIODE Outdoor. This is because $\mathcal{L}_{tp-si}$ provides more comprehensive guidance on pixels without ground truth, thus enhancing the overall generalization ability for zero-shot evaluation. Further discarding rescale maps results in a significant decline in all metrics, demonstrating the importance of using pixel-level maps to do rescaling.

\begin{table}[t]
  \fontsize{14}{12}\selectfont
  \centering
  \begin{minipage}{0.46\textwidth}
    \captionof{table}{Ablation study on the main proposed modules.}
    \centering
    \resizebox{0.95\textwidth}{!}{
    \begin{tabular}{ccc|cccccc}
    \toprule
    \multirow{2}{*}{Rescale Maps} & \multirow{2}{*}{$\mathcal{L}_{tp-si}$} & \multirow{2}{*}{$\mathcal{L}_{soc}$} & \multicolumn{2}{c}{NYUv2} &  \multicolumn{2}{c}{iBims-1} &  \multicolumn{2}{c}{DIODE Outdoor} \\
     & & & $Abs Rel\downarrow$ & $\delta_1 \uparrow$ & $Abs Rel\downarrow$ & $\delta_1 \uparrow$ & $Abs Rel\downarrow$ & $\delta_1 \uparrow$ \\ \midrule
     \xmark & \xmark & \xmark & 0.118 & 0.902 & 0.202 & 0.566 & 0.761 & 0.225 \\
    \cmark & \xmark & \xmark & 0.084 & 0.946 & 0.173 & 0.657 & 0.702 & 0.237 \\
    \cmark & \cmark & \xmark & 0.085 & 0.945 & 0.160 & 0.708 & 0.687 & 0.243 \\
    \cmark & \cmark & \cmark & 0.082 & 0.954 & 0.154 & 0.736 & 0.673 & 0.274 \\
    \bottomrule
  \end{tabular}}
    
    \label{tab:ablation_module}
  \end{minipage}
  \hfill
  \begin{minipage}{0.46\textwidth}
  \captionof{table}{Ablation study on text and image information.}
    \centering
    \resizebox{0.95\textwidth}{!}{
    \begin{tabular}{cc|cccccc}
\toprule
    \multirow{2}{*}{Text} & \multirow{2}{*}{Image} & \multicolumn{2}{c}{NYUv2} &  \multicolumn{2}{c}{iBims-1} &  \multicolumn{2}{c}{DIODE Outdoor} \\
     & & $Abs Rel\downarrow$ & $\delta_1 \uparrow$ & $Abs Rel\downarrow$ & $\delta_1 \uparrow$ & $Abs Rel\downarrow$ & $\delta_1 \uparrow$ \\ \midrule
    \cmark & \xmark & 0.114 & 0.909 & 0.193 & 0.621 & 0.745 & 0.231 \\
    \xmark & \cmark & 0.085 & 0.945 & 0.164 & 0.704 & 0.694 & 0.259 \\
    \cmark & \cmark &  0.082 & 0.954 & 0.154 & 0.736 & 0.673 & 0.274 \\
    \bottomrule
\end{tabular}}
    
    \label{tab:ablation_text_image}
  \end{minipage}
\end{table}

\paragraph{Incorporating text and image information.}
We further carry out an ablation study on the effects of text and image information for relative to metric depth transfer. The results are presented in Table~\ref{tab:ablation_text_image}. Text descriptions cannot capture information about depth distribution or layouts, resulting in inaccurate scale estimation when using text information only. Using only image information improves performance substantially, but it is still not as good as using both text and image. The information from the two modalities can complement each other, ensuring the model's perception of both global and local scales.
\section{Conclusion}
\label{sec:conclusion}

This paper proposes TR2M, a generalizable framework for transferring monocular relative depth to metric depth. TR2M takes a text description and an image, predicting two rescaling maps for pixel-wise transformation. A cross-modality attention module fuses features from both modalities. To leverage relative depth effectively, we align it with ground truth to generate pseudo metric depth, filtering confident samples for supervision. We further introduce scale-oriented contrastive learning to enhance scale perception by modeling depth distribution. Consequently, TR2M demonstrates strong zero-shot metric depth estimation capability across diverse domains using a single lightweight network.

\section*{Acknowledgements.}
This work was supported by Hong Kong RGC CRF C4026-21G,  RIF R4020-22, GRF 14211420, 14216020 \& 14203323).

{
    \small
    \bibliographystyle{ieeenat_fullname}
    \bibliography{main}

\begin{thebibliography}{76}
\providecommand{\natexlab}[1]{#1}
\providecommand{\url}[1]{\texttt{#1}}
\expandafter\ifx\csname urlstyle\endcsname\relax
  \providecommand{\doi}[1]{doi: #1}\else
  \providecommand{\doi}{doi: \begingroup \urlstyle{rm}\Url}\fi

\bibitem[Antequera et~al.(2020)Antequera, Gargallo, Hofinger, Bulo, Kuang, and Kontschieder]{antequera2020mapillary}
Manuel~L{\'o}pez Antequera, Pau Gargallo, Markus Hofinger, Samuel~Rota Bulo, Yubin Kuang, and Peter Kontschieder.
\newblock Mapillary planet-scale depth dataset.
\newblock In \emph{European Conference on Computer Vision}, pages 589--604. Springer, 2020.

\bibitem[Auty and Mikolajczyk(2023)]{auty2023learning}
Dylan Auty and Krystian Mikolajczyk.
\newblock Learning to prompt clip for monocular depth estimation: Exploring the limits of human language.
\newblock In \emph{Proceedings of the IEEE/CVF International Conference on Computer Vision}, pages 2039--2047, 2023.

\bibitem[Bhat et~al.(2021)Bhat, Alhashim, and Wonka]{bhat2021adabins}
Shariq~Farooq Bhat, Ibraheem Alhashim, and Peter Wonka.
\newblock Adabins: Depth estimation using adaptive bins.
\newblock In \emph{Proceedings of the IEEE/CVF conference on computer vision and pattern recognition}, pages 4009--4018, 2021.

\bibitem[Bhat et~al.(2022)Bhat, Alhashim, and Wonka]{bhat2022localbins}
Shariq~Farooq Bhat, Ibraheem Alhashim, and Peter Wonka.
\newblock Localbins: Improving depth estimation by learning local distributions.
\newblock In \emph{European Conference on Computer Vision}, pages 480--496. Springer, 2022.

\bibitem[Bhat et~al.(2023)Bhat, Birkl, Wofk, Wonka, and M{\"u}ller]{bhat2023zoedepth}
Shariq~Farooq Bhat, Reiner Birkl, Diana Wofk, Peter Wonka, and Matthias M{\"u}ller.
\newblock Zoedepth: Zero-shot transfer by combining relative and metric depth.
\newblock \emph{arXiv preprint arXiv:2302.12288}, 2023.

\bibitem[Bobrow et~al.(2023)Bobrow, Golhar, Vijayan, Akshintala, Garcia, and Durr]{bobrow2023}
Taylor~L Bobrow, Mayank Golhar, Rohan Vijayan, Venkata~S Akshintala, Juan~R Garcia, and Nicholas~J Durr.
\newblock Colonoscopy 3d video dataset with paired depth from 2d-3d registration.
\newblock \emph{Medical Image Analysis}, page 102956, 2023.

\bibitem[Cai et~al.(2025)Cai, Yeh, Xu, Liu, Meyer, Lei, Zhao, Li, Chandra, and Shi]{cai2025depthlm}
Zhipeng Cai, Ching-Feng Yeh, Hu Xu, Zhuang Liu, Gregory Meyer, Xinjie Lei, Changsheng Zhao, Shang-Wen Li, Vikas Chandra, and Yangyang Shi.
\newblock Depthlm: Metric depth from vision language models.
\newblock \emph{arXiv preprint arXiv:2509.25413}, 2025.

\bibitem[Caron et~al.(2021)Caron, Touvron, Misra, J{\'e}gou, Mairal, Bojanowski, and Joulin]{caron2021emerging}
Mathilde Caron, Hugo Touvron, Ishan Misra, Herv{\'e} J{\'e}gou, Julien Mairal, Piotr Bojanowski, and Armand Joulin.
\newblock Emerging properties in self-supervised vision transformers.
\newblock In \emph{Proceedings of the IEEE/CVF international conference on computer vision}, pages 9650--9660, 2021.

\bibitem[Chen et~al.(2015)Chen, Seff, Kornhauser, and Xiao]{chen2015deepdriving}
Chenyi Chen, Ari Seff, Alain Kornhauser, and Jianxiong Xiao.
\newblock Deepdriving: Learning affordance for direct perception in autonomous driving.
\newblock In \emph{Proceedings of the IEEE international conference on computer vision}, pages 2722--2730, 2015.

\bibitem[Chen et~al.(2020)Chen, Fan, Girshick, and He]{chen2020improved}
Xinlei Chen, Haoqi Fan, Ross Girshick, and Kaiming He.
\newblock Improved baselines with momentum contrastive learning.
\newblock \emph{arXiv preprint arXiv:2003.04297}, 2020.

\bibitem[Dong et~al.(2022)Dong, Garratt, Anavatti, and Abbass]{dong2022towards}
Xingshuai Dong, Matthew~A Garratt, Sreenatha~G Anavatti, and Hussein~A Abbass.
\newblock Towards real-time monocular depth estimation for robotics: A survey.
\newblock \emph{IEEE Transactions on Intelligent Transportation Systems}, 23\penalty0 (10):\penalty0 16940--16961, 2022.

\bibitem[Du et~al.(2020)Du, Turner, Dzitsiuk, Prasso, Duarte, Dourgarian, Afonso, Pascoal, Gladstone, Cruces, et~al.]{du2020depthlab}
Ruofei Du, Eric Turner, Maksym Dzitsiuk, Luca Prasso, Ivo Duarte, Jason Dourgarian, Joao Afonso, Jose Pascoal, Josh Gladstone, Nuno Cruces, et~al.
\newblock Depthlab: Real-time 3d interaction with depth maps for mobile augmented reality.
\newblock In \emph{Proceedings of the 33rd Annual ACM Symposium on User Interface Software and Technology}, pages 829--843, 2020.

\bibitem[Eigen et~al.(2014)Eigen, Puhrsch, and Fergus]{eigen2014depth}
David Eigen, Christian Puhrsch, and Rob Fergus.
\newblock Depth map prediction from a single image using a multi-scale deep network.
\newblock \emph{Advances in neural information processing systems}, 27, 2014.

\bibitem[El~Jamiy and Marsh(2019)]{el2019survey}
Fatima El~Jamiy and Ronald Marsh.
\newblock Survey on depth perception in head mounted displays: distance estimation in virtual reality, augmented reality, and mixed reality.
\newblock \emph{IET Image Processing}, 13\penalty0 (5):\penalty0 707--712, 2019.

\bibitem[Facil et~al.(2019)Facil, Ummenhofer, Zhou, Montesano, Brox, and Civera]{facil2019cam}
Jose~M Facil, Benjamin Ummenhofer, Huizhong Zhou, Luis Montesano, Thomas Brox, and Javier Civera.
\newblock Cam-convs: Camera-aware multi-scale convolutions for single-view depth.
\newblock In \emph{Proceedings of the IEEE/CVF Conference on Computer Vision and Pattern Recognition}, pages 11826--11835, 2019.

\bibitem[Fan et~al.(2023)Fan, Poggi, and Mattoccia]{fan2023contrastive}
Rizhao Fan, Matteo Poggi, and Stefano Mattoccia.
\newblock Contrastive learning for depth prediction.
\newblock In \emph{Proceedings of the IEEE/CVF conference on computer vision and pattern recognition}, pages 3226--3237, 2023.

\bibitem[Fu et~al.(2018)Fu, Gong, Wang, Batmanghelich, and Tao]{fu2018deep}
Huan Fu, Mingming Gong, Chaohui Wang, Kayhan Batmanghelich, and Dacheng Tao.
\newblock Deep ordinal regression network for monocular depth estimation.
\newblock In \emph{Proceedings of the IEEE conference on computer vision and pattern recognition}, pages 2002--2011, 2018.

\bibitem[Geiger et~al.(2012)Geiger, Lenz, and Urtasun]{geiger2012we}
Andreas Geiger, Philip Lenz, and Raquel Urtasun.
\newblock Are we ready for autonomous driving? the kitti vision benchmark suite.
\newblock In \emph{2012 IEEE conference on computer vision and pattern recognition}, pages 3354--3361. IEEE, 2012.

\bibitem[Godard et~al.(2017)Godard, Mac~Aodha, and Brostow]{godard2017unsupervised}
Cl{\'e}ment Godard, Oisin Mac~Aodha, and Gabriel~J Brostow.
\newblock Unsupervised monocular depth estimation with left-right consistency.
\newblock In \emph{Proceedings of the IEEE conference on computer vision and pattern recognition}, pages 270--279, 2017.

\bibitem[Guizilini et~al.(2023)Guizilini, Vasiljevic, Chen, Ambruș, and Gaidon]{guizilini2023towards}
Vitor Guizilini, Igor Vasiljevic, Dian Chen, Rareș Ambruș, and Adrien Gaidon.
\newblock Towards zero-shot scale-aware monocular depth estimation.
\newblock In \emph{Proceedings of the IEEE/CVF International Conference on Computer Vision}, pages 9233--9243, 2023.

\bibitem[Han et~al.(2023)Han, Yin, and Shen]{Han_2023_ICCV}
Wencheng Han, Junbo Yin, and Jianbing Shen.
\newblock Self-supervised monocular depth estimation by direction-aware cumulative convolution network.
\newblock In \emph{Proceedings of the IEEE/CVF International Conference on Computer Vision (ICCV)}, pages 8613--8623, 2023.

\bibitem[He et~al.(2020)He, Fan, Wu, Xie, and Girshick]{he2020momentum}
Kaiming He, Haoqi Fan, Yuxin Wu, Saining Xie, and Ross Girshick.
\newblock Momentum contrast for unsupervised visual representation learning.
\newblock In \emph{Proceedings of the IEEE/CVF conference on computer vision and pattern recognition}, pages 9729--9738, 2020.

\bibitem[Hu et~al.(2024{\natexlab{a}})Hu, Yin, Zhang, Cai, Long, Chen, Wang, Yu, Shen, and Shen]{hu2024metric3d}
Mu Hu, Wei Yin, Chi Zhang, Zhipeng Cai, Xiaoxiao Long, Hao Chen, Kaixuan Wang, Gang Yu, Chunhua Shen, and Shaojie Shen.
\newblock Metric3d v2: A versatile monocular geometric foundation model for zero-shot metric depth and surface normal estimation.
\newblock \emph{IEEE Transactions on Pattern Analysis and Machine Intelligence}, 2024{\natexlab{a}}.

\bibitem[Hu et~al.(2024{\natexlab{b}})Hu, Zhang, Zhang, Hai, Yu, and He]{hu2024learning}
Xueting Hu, Ce Zhang, Yi Zhang, Bowen Hai, Ke Yu, and Zhihai He.
\newblock Learning to adapt clip for few-shot monocular depth estimation.
\newblock In \emph{Proceedings of the IEEE/CVF Winter Conference on Applications of Computer Vision}, pages 5594--5603, 2024{\natexlab{b}}.

\bibitem[Huang et~al.(2024)Huang, Cui, Bai, Guo, Xu, Islam, and Ren]{huang2024endo}
Yiming Huang, Beilei Cui, Long Bai, Ziqi Guo, Mengya Xu, Mobarakol Islam, and Hongliang Ren.
\newblock Endo-4dgs: Endoscopic monocular scene reconstruction with 4d gaussian splatting.
\newblock In \emph{International Conference on Medical Image Computing and Computer-Assisted Intervention}, pages 197--207. Springer, 2024.

\bibitem[Hunter(1986)]{hunter1986exponentially}
J~Stuart Hunter.
\newblock The exponentially weighted moving average.
\newblock \emph{Journal of quality technology}, 18\penalty0 (4):\penalty0 203--210, 1986.

\bibitem[Jin et~al.(2022)Jin, Yu, Chen, Zhao, Heng, and Stoyanov]{jin2022exploring}
Yueming Jin, Yang Yu, Cheng Chen, Zixu Zhao, Pheng-Ann Heng, and Danail Stoyanov.
\newblock Exploring intra-and inter-video relation for surgical semantic scene segmentation.
\newblock \emph{IEEE Transactions on Medical Imaging}, 41\penalty0 (11):\penalty0 2991--3002, 2022.

\bibitem[Ke et~al.(2024)Ke, Obukhov, Huang, Metzger, Daudt, and Schindler]{ke2024repurposing}
Bingxin Ke, Anton Obukhov, Shengyu Huang, Nando Metzger, Rodrigo~Caye Daudt, and Konrad Schindler.
\newblock Repurposing diffusion-based image generators for monocular depth estimation.
\newblock In \emph{Proceedings of the IEEE/CVF Conference on Computer Vision and Pattern Recognition}, pages 9492--9502, 2024.

\bibitem[Koch et~al.(2018)Koch, Liebel, Fraundorfer, and Korner]{koch2018evaluation}
Tobias Koch, Lukas Liebel, Friedrich Fraundorfer, and Marco Korner.
\newblock Evaluation of cnn-based single-image depth estimation methods.
\newblock In \emph{Proceedings of the European Conference on Computer Vision (ECCV) Workshops}, pages 0--0, 2018.

\bibitem[Laina et~al.(2016)Laina, Rupprecht, Belagiannis, Tombari, and Navab]{laina2016deeper}
Iro Laina, Christian Rupprecht, Vasileios Belagiannis, Federico Tombari, and Nassir Navab.
\newblock Deeper depth prediction with fully convolutional residual networks.
\newblock In \emph{2016 Fourth international conference on 3D vision (3DV)}, pages 239--248. IEEE, 2016.

\bibitem[Li et~al.(2024)Li, Li, Wang, Wang, Yang, and Zou]{li2024region}
Meixuan Li, Tianyu Li, Guoqing Wang, Peng Wang, Yang Yang, and Jie Zou.
\newblock Region-aware distribution contrast: A novel approach to multi-task partially supervised learning.
\newblock In \emph{European Conference on Computer Vision}, pages 234--251. Springer, 2024.

\bibitem[Liu et~al.(2024)Liu, Li, Li, and Lee]{liu2024improved}
Haotian Liu, Chunyuan Li, Yuheng Li, and Yong~Jae Lee.
\newblock Improved baselines with visual instruction tuning.
\newblock In \emph{Proceedings of the IEEE/CVF Conference on Computer Vision and Pattern Recognition}, pages 26296--26306, 2024.

\bibitem[Liu et~al.(2019)Liu, Sinha, Ishii, Hager, Reiter, Taylor, and Unberath]{liu2019dense}
Xingtong Liu, Ayushi Sinha, Masaru Ishii, Gregory~D Hager, Austin Reiter, Russell~H Taylor, and Mathias Unberath.
\newblock Dense depth estimation in monocular endoscopy with self-supervised learning methods.
\newblock \emph{IEEE transactions on medical imaging}, 39\penalty0 (5):\penalty0 1438--1447, 2019.

\bibitem[Loshchilov and Hutter(2017)]{loshchilov2017decoupled}
Ilya Loshchilov and Frank Hutter.
\newblock Decoupled weight decay regularization.
\newblock \emph{arXiv preprint arXiv:1711.05101}, 2017.

\bibitem[Oquab et~al.(2023)Oquab, Darcet, Moutakanni, Vo, Szafraniec, Khalidov, Fernandez, Haziza, Massa, El-Nouby, et~al.]{oquab2023dinov2}
Maxime Oquab, Timoth{\'e}e Darcet, Th{\'e}o Moutakanni, Huy Vo, Marc Szafraniec, Vasil Khalidov, Pierre Fernandez, Daniel Haziza, Francisco Massa, Alaaeldin El-Nouby, et~al.
\newblock Dinov2: Learning robust visual features without supervision.
\newblock \emph{arXiv preprint arXiv:2304.07193}, 2023.

\bibitem[Patni et~al.(2024)Patni, Agarwal, and Arora]{patni2024ecodepth}
Suraj Patni, Aradhye Agarwal, and Chetan Arora.
\newblock Ecodepth: Effective conditioning of diffusion models for monocular depth estimation.
\newblock In \emph{Proceedings of the IEEE/CVF Conference on Computer Vision and Pattern Recognition}, pages 28285--28295, 2024.

\bibitem[Peng et~al.(2020)Peng, Pan, Liu, and Sun]{peng2020ida}
Wanli Peng, Hao Pan, He Liu, and Yi Sun.
\newblock Ida-3d: Instance-depth-aware 3d object detection from stereo vision for autonomous driving.
\newblock In \emph{Proceedings of the IEEE/CVF conference on computer vision and pattern recognition}, pages 13015--13024, 2020.

\bibitem[Piccinelli et~al.(2023)Piccinelli, Sakaridis, and Yu]{piccinelli2023idisc}
Luigi Piccinelli, Christos Sakaridis, and Fisher Yu.
\newblock idisc: Internal discretization for monocular depth estimation.
\newblock In \emph{Proceedings of the IEEE/CVF Conference on Computer Vision and Pattern Recognition}, pages 21477--21487, 2023.

\bibitem[Piccinelli et~al.(2024)Piccinelli, Yang, Sakaridis, Segu, Li, Van~Gool, and Yu]{piccinelli2024unidepth}
Luigi Piccinelli, Yung-Hsu Yang, Christos Sakaridis, Mattia Segu, Siyuan Li, Luc Van~Gool, and Fisher Yu.
\newblock Unidepth: Universal monocular metric depth estimation.
\newblock In \emph{Proceedings of the IEEE/CVF Conference on Computer Vision and Pattern Recognition}, pages 10106--10116, 2024.

\bibitem[Piccinelli et~al.(2025)Piccinelli, Sakaridis, Segu, Yang, Li, Abbeloos, and Van~Gool]{piccinelli2025unik3d}
Luigi Piccinelli, Christos Sakaridis, Mattia Segu, Yung-Hsu Yang, Siyuan Li, Wim Abbeloos, and Luc Van~Gool.
\newblock Unik3d: Universal camera monocular 3d estimation.
\newblock In \emph{Proceedings of the Computer Vision and Pattern Recognition Conference}, pages 1028--1039, 2025.

\bibitem[Radford et~al.(2021)Radford, Kim, Hallacy, Ramesh, Goh, Agarwal, Sastry, Askell, Mishkin, Clark, et~al.]{radford2021learning}
Alec Radford, Jong~Wook Kim, Chris Hallacy, Aditya Ramesh, Gabriel Goh, Sandhini Agarwal, Girish Sastry, Amanda Askell, Pamela Mishkin, Jack Clark, et~al.
\newblock Learning transferable visual models from natural language supervision.
\newblock In \emph{International conference on machine learning}, pages 8748--8763. PmLR, 2021.

\bibitem[Ranftl et~al.(2021{\natexlab{a}})Ranftl, Bochkovskiy, and Koltun]{Ranftl2021}
Ren\'{e} Ranftl, Alexey Bochkovskiy, and Vladlen Koltun.
\newblock Vision transformers for dense prediction.
\newblock \emph{ICCV}, 2021{\natexlab{a}}.

\bibitem[Ranftl et~al.(2021{\natexlab{b}})Ranftl, Bochkovskiy, and Koltun]{ranftl2021vision}
Ren{\'e} Ranftl, Alexey Bochkovskiy, and Vladlen Koltun.
\newblock Vision transformers for dense prediction.
\newblock In \emph{Proceedings of the IEEE/CVF international conference on computer vision}, pages 12179--12188, 2021{\natexlab{b}}.

\bibitem[Ranftl et~al.(2022)Ranftl, Lasinger, Hafner, Schindler, and Koltun]{Ranftl2022}
Ren\'{e} Ranftl, Katrin Lasinger, David Hafner, Konrad Schindler, and Vladlen Koltun.
\newblock Towards robust monocular depth estimation: Mixing datasets for zero-shot cross-dataset transfer.
\newblock \emph{IEEE Transactions on Pattern Analysis and Machine Intelligence}, 44\penalty0 (3), 2022.

\bibitem[Rau et~al.(2023)Rau, Bhattarai, Agapito, and Stoyanov]{rau2023bimodal}
Anita Rau, Binod Bhattarai, Lourdes Agapito, and Danail Stoyanov.
\newblock Bimodal camera pose prediction for endoscopy.
\newblock \emph{IEEE Transactions on Medical Robotics and Bionics}, 5\penalty0 (4):\penalty0 978--989, 2023.

\bibitem[Roberts et~al.(2021)Roberts, Ramapuram, Ranjan, Kumar, Bautista, Paczan, Webb, and Susskind]{roberts2021hypersim}
Mike Roberts, Jason Ramapuram, Anurag Ranjan, Atulit Kumar, Miguel~Angel Bautista, Nathan Paczan, Russ Webb, and Joshua~M Susskind.
\newblock Hypersim: A photorealistic synthetic dataset for holistic indoor scene understanding.
\newblock In \emph{Proceedings of the IEEE/CVF international conference on computer vision}, pages 10912--10922, 2021.

\bibitem[Rodr{\'\i}guez-Puigvert et~al.(2023)Rodr{\'\i}guez-Puigvert, Batlle, Montiel, Martinez-Cantin, Fua, Tard\'os, and Civera]{Rodriguez-Puigvert_2023_ICCV}
Javier Rodr{\'\i}guez-Puigvert, V{\'\i}ctor~M. Batlle, J.M.M. Montiel, Ruben Martinez-Cantin, Pascal Fua, Juan~D. Tard\'os, and Javier Civera.
\newblock Lightdepth: Single-view depth self-supervision from illumination decline.
\newblock In \emph{Proceedings of the IEEE/CVF International Conference on Computer Vision (ICCV)}, pages 21273--21283, 2023.

\bibitem[Shao et~al.(2023)Shao, Pei, Chen, Wu, and Li]{Shao_2023_ICCV}
Shuwei Shao, Zhongcai Pei, Weihai Chen, Xingming Wu, and Zhengguo Li.
\newblock Nddepth: Normal-distance assisted monocular depth estimation.
\newblock In \emph{Proceedings of the IEEE/CVF International Conference on Computer Vision (ICCV)}, pages 7931--7940, 2023.

\bibitem[Silberman et~al.(2012)Silberman, Hoiem, Kohli, and Fergus]{silberman2012indoor}
Nathan Silberman, Derek Hoiem, Pushmeet Kohli, and Rob Fergus.
\newblock Indoor segmentation and support inference from rgbd images.
\newblock In \emph{Computer Vision--ECCV 2012: 12th European Conference on Computer Vision, Florence, Italy, October 7-13, 2012, Proceedings, Part V 12}, pages 746--760. Springer, 2012.

\bibitem[Song et~al.(2015)Song, Lichtenberg, and Xiao]{song2015sun}
Shuran Song, Samuel~P Lichtenberg, and Jianxiong Xiao.
\newblock Sun rgb-d: A rgb-d scene understanding benchmark suite.
\newblock In \emph{Proceedings of the IEEE conference on computer vision and pattern recognition}, pages 567--576, 2015.

\bibitem[Vasiljevic et~al.(2019)Vasiljevic, Kolkin, Zhang, Luo, Wang, Dai, Daniele, Mostajabi, Basart, Walter, et~al.]{vasiljevic2019diode}
Igor Vasiljevic, Nick Kolkin, Shanyi Zhang, Ruotian Luo, Haochen Wang, Falcon~Z Dai, Andrea~F Daniele, Mohammadreza Mostajabi, Steven Basart, Matthew~R Walter, et~al.
\newblock Diode: A dense indoor and outdoor depth dataset.
\newblock \emph{arXiv preprint arXiv:1908.00463}, 2019.

\bibitem[Wang et~al.(2021{\natexlab{a}})Wang, Zhou, Yu, Dai, Konukoglu, and Van~Gool]{wang2021exploring}
Wenguan Wang, Tianfei Zhou, Fisher Yu, Jifeng Dai, Ender Konukoglu, and Luc Van~Gool.
\newblock Exploring cross-image pixel contrast for semantic segmentation.
\newblock In \emph{Proceedings of the IEEE/CVF international conference on computer vision}, pages 7303--7313, 2021{\natexlab{a}}.

\bibitem[Wang et~al.(2021{\natexlab{b}})Wang, Zhang, Shen, Kong, and Li]{wang2021dense}
Xinlong Wang, Rufeng Zhang, Chunhua Shen, Tao Kong, and Lei Li.
\newblock Dense contrastive learning for self-supervised visual pre-training.
\newblock In \emph{Proceedings of the IEEE/CVF conference on computer vision and pattern recognition}, pages 3024--3033, 2021{\natexlab{b}}.

\bibitem[Wofk et~al.(2019)Wofk, Ma, Yang, Karaman, and Sze]{wofk2019fastdepth}
Diana Wofk, Fangchang Ma, Tien-Ju Yang, Sertac Karaman, and Vivienne Sze.
\newblock Fastdepth: Fast monocular depth estimation on embedded systems.
\newblock In \emph{2019 International Conference on Robotics and Automation (ICRA)}, pages 6101--6108. IEEE, 2019.

\bibitem[Wong et~al.(2020{\natexlab{a}})Wong, Cicek, and Soatto]{wong2020targeted}
Alex Wong, Safa Cicek, and Stefano Soatto.
\newblock Targeted adversarial perturbations for monocular depth prediction.
\newblock \emph{Advances in neural information processing systems}, 33:\penalty0 8486--8497, 2020{\natexlab{a}}.

\bibitem[Wong et~al.(2020{\natexlab{b}})Wong, Fei, Tsuei, and Soatto]{wong2020unsupervised}
Alex Wong, Xiaohan Fei, Stephanie Tsuei, and Stefano Soatto.
\newblock Unsupervised depth completion from visual inertial odometry.
\newblock \emph{IEEE Robotics and Automation Letters}, 5\penalty0 (2):\penalty0 1899--1906, 2020{\natexlab{b}}.

\bibitem[Wu et~al.(2022)Wu, Wang, Hall, Neumann, and Su]{wu2022toward}
Cho-Ying Wu, Jialiang Wang, Michael Hall, Ulrich Neumann, and Shuochen Su.
\newblock Toward practical monocular indoor depth estimation.
\newblock In \emph{CVPR}, 2022.

\bibitem[Wu et~al.(2018)Wu, Xiong, Yu, and Lin]{wu2018unsupervised}
Zhirong Wu, Yuanjun Xiong, Stella~X Yu, and Dahua Lin.
\newblock Unsupervised feature learning via non-parametric instance discrimination.
\newblock In \emph{Proceedings of the IEEE conference on computer vision and pattern recognition}, pages 3733--3742, 2018.

\bibitem[Yang et~al.(2024{\natexlab{a}})Yang, Feng, Chen, Park, Wang, Dou, Zeng, Chen, Gangopadhyay, Owens, et~al.]{yang2024binding}
Fengyu Yang, Chao Feng, Ziyang Chen, Hyoungseob Park, Daniel Wang, Yiming Dou, Ziyao Zeng, Xien Chen, Rit Gangopadhyay, Andrew Owens, et~al.
\newblock Binding touch to everything: Learning unified multimodal tactile representations.
\newblock In \emph{Proceedings of the IEEE/CVF Conference on Computer Vision and Pattern Recognition}, pages 26340--26353, 2024{\natexlab{a}}.

\bibitem[Yang et~al.(2021)Yang, Tang, Ding, Sebe, and Ricci]{yang2021transformer}
Guanglei Yang, Hao Tang, Mingli Ding, Nicu Sebe, and Elisa Ricci.
\newblock Transformer-based attention networks for continuous pixel-wise prediction.
\newblock In \emph{Proceedings of the IEEE/CVF International Conference on Computer vision}, pages 16269--16279, 2021.

\bibitem[Yang et~al.(2024{\natexlab{b}})Yang, Kang, Huang, Xu, Feng, and Zhao]{yang2024depth}
Lihe Yang, Bingyi Kang, Zilong Huang, Xiaogang Xu, Jiashi Feng, and Hengshuang Zhao.
\newblock Depth anything: Unleashing the power of large-scale unlabeled data.
\newblock In \emph{Proceedings of the IEEE/CVF Conference on Computer Vision and Pattern Recognition}, pages 10371--10381, 2024{\natexlab{b}}.

\bibitem[Yang et~al.(2025)Yang, Kang, Huang, Zhao, Xu, Feng, and Zhao]{yang2025depth}
Lihe Yang, Bingyi Kang, Zilong Huang, Zhen Zhao, Xiaogang Xu, Jiashi Feng, and Hengshuang Zhao.
\newblock Depth anything v2.
\newblock \emph{Advances in Neural Information Processing Systems}, 37:\penalty0 21875--21911, 2025.

\bibitem[Yang et~al.(2023)Yang, Ma, Ji, and Ren]{Yang_2023_ICCV}
Xiaodong Yang, Zhuang Ma, Zhiyu Ji, and Zhe Ren.
\newblock Gedepth: Ground embedding for monocular depth estimation.
\newblock In \emph{Proceedings of the IEEE/CVF International Conference on Computer Vision (ICCV)}, pages 12719--12727, 2023.

\bibitem[Yin et~al.(2023)Yin, Zhang, Chen, Cai, Yu, Wang, Chen, and Shen]{yin2023metric3d}
Wei Yin, Chi Zhang, Hao Chen, Zhipeng Cai, Gang Yu, Kaixuan Wang, Xiaozhi Chen, and Chunhua Shen.
\newblock Metric3d: Towards zero-shot metric 3d prediction from a single image.
\newblock In \emph{Proceedings of the IEEE/CVF International Conference on Computer Vision}, pages 9043--9053, 2023.

\bibitem[You et~al.(2024)You, Mint, Dai, Sekhon, Staib, and Duncan]{you2024calibrating}
Chenyu You, Yifei Mint, Weicheng Dai, Jasjeet~S Sekhon, Lawrence Staib, and James~S Duncan.
\newblock Calibrating multi-modal representations: A pursuit of group robustness without annotations.
\newblock In \emph{2024 IEEE/CVF Conference on Computer Vision and Pattern Recognition (CVPR)}, pages 26140--26150. IEEE, 2024.

\bibitem[Yuan et~al.(2022)Yuan, Gu, Dai, Zhu, and Tan]{yuan2022neural}
Weihao Yuan, Xiaodong Gu, Zuozhuo Dai, Siyu Zhu, and Ping Tan.
\newblock Neural window fully-connected crfs for monocular depth estimation.
\newblock In \emph{Proceedings of the IEEE/CVF conference on computer vision and pattern recognition}, pages 3916--3925, 2022.

\bibitem[Zeng et~al.(2024{\natexlab{a}})Zeng, Wang, Yang, Park, Soatto, Lao, and Wong]{zeng2024wordepth}
Ziyao Zeng, Daniel Wang, Fengyu Yang, Hyoungseob Park, Stefano Soatto, Dong Lao, and Alex Wong.
\newblock Wordepth: Variational language prior for monocular depth estimation.
\newblock In \emph{Proceedings of the IEEE/CVF Conference on Computer Vision and Pattern Recognition}, pages 9708--9719, 2024{\natexlab{a}}.

\bibitem[Zeng et~al.(2024{\natexlab{b}})Zeng, Wu, Park, Wang, Yang, Soatto, Lao, Hong, and Wong]{zengrsa}
Ziyao Zeng, Yangchao Wu, Hyoungseob Park, Daniel Wang, Fengyu Yang, Stefano Soatto, Dong Lao, Byung-Woo Hong, and Alex Wong.
\newblock Rsa: Resolving scale ambiguities in monocular depth estimators through language descriptions.
\newblock In \emph{The Thirty-eighth Annual Conference on Neural Information Processing Systems}, 2024{\natexlab{b}}.

\bibitem[Zhang et~al.(2022{\natexlab{a}})Zhang, Guo, Zhang, Li, Miao, Cui, Qiao, Gao, and Li]{zhang2022pointclip}
Renrui Zhang, Ziyu Guo, Wei Zhang, Kunchang Li, Xupeng Miao, Bin Cui, Yu Qiao, Peng Gao, and Hongsheng Li.
\newblock Pointclip: Point cloud understanding by clip.
\newblock In \emph{Proceedings of the IEEE/CVF conference on computer vision and pattern recognition}, pages 8552--8562, 2022{\natexlab{a}}.

\bibitem[Zhang et~al.(2022{\natexlab{b}})Zhang, Zeng, Guo, and Li]{zhang2022can}
Renrui Zhang, Ziyao Zeng, Ziyu Guo, and Yafeng Li.
\newblock Can language understand depth?
\newblock In \emph{Proceedings of the 30th ACM International Conference on Multimedia}, pages 6868--6874, 2022{\natexlab{b}}.

\bibitem[Zhang et~al.(2025)Zhang, Liu, Li, He, Qi, Wang, Zhao, Yu, Zeng, and Jin]{zhang2025hybrid}
Wenyao Zhang, Hongsi Liu, Bohan Li, Jiawei He, Zekun Qi, Yunnan Wang, Shengyang Zhao, Xinqiang Yu, Wenjun Zeng, and Xin Jin.
\newblock Hybrid-grained feature aggregation with coarse-to-fine language guidance for self-supervised monocular depth estimation.
\newblock In \emph{Proceedings of the IEEE/CVF International Conference on Computer Vision}, pages 6678--6692, 2025.

\bibitem[Zhao et~al.(2023)Zhao, Poggi, Tosi, Zhou, Sun, Tang, and Mattoccia]{Zhao_2023_ICCV}
Chaoqiang Zhao, Matteo Poggi, Fabio Tosi, Lei Zhou, Qiyu Sun, Yang Tang, and Stefano Mattoccia.
\newblock Gasmono: Geometry-aided self-supervised monocular depth estimation for indoor scenes.
\newblock In \emph{Proceedings of the IEEE/CVF International Conference on Computer Vision (ICCV)}, pages 16209--16220, 2023.

\bibitem[Zhao et~al.(2021)Zhao, Vemulapalli, Mansfield, Gong, Green, Shapira, and Wu]{zhao2021contrastive}
Xiangyun Zhao, Raviteja Vemulapalli, Philip~Andrew Mansfield, Boqing Gong, Bradley Green, Lior Shapira, and Ying Wu.
\newblock Contrastive learning for label efficient semantic segmentation.
\newblock In \emph{Proceedings of the IEEE/CVF International Conference on Computer Vision}, pages 10623--10633, 2021.

\bibitem[Zhao et~al.(2024)Zhao, Bian, Chen, Ji, Qu, Lin, Yu, Li, Chen, Shen, et~al.]{zhao2024metric}
Yizhou Zhao, Hengwei Bian, Kaihua Chen, Pengliang Ji, Liao Qu, Shao-yu Lin, Weichen Yu, Haoran Li, Hao Chen, Jun Shen, et~al.
\newblock Metric from human: Zero-shot monocular metric depth estimation via test-time adaptation.
\newblock In \emph{The Thirty-eighth Annual Conference on Neural Information Processing Systems}, 2024.

\bibitem[Zhu et~al.(2024)Zhu, Wang, Song, Liu, Zhang, and Zhang]{zhu2024scale}
Ruijie Zhu, Chuxin Wang, Ziyang Song, Li Liu, Tianzhu Zhang, and Yongdong Zhang.
\newblock Scaledepth: Decomposing metric depth estimation into scale prediction and relative depth estimation.
\newblock \emph{arXiv preprint arXiv:2407.08187}, 2024.

\bibitem[Zhu et~al.(2023)Zhu, Zhang, He, Guo, Zeng, Qin, Zhang, and Gao]{zhu2023pointclip}
Xiangyang Zhu, Renrui Zhang, Bowei He, Ziyu Guo, Ziyao Zeng, Zipeng Qin, Shanghang Zhang, and Peng Gao.
\newblock Pointclip v2: Prompting clip and gpt for powerful 3d open-world learning.
\newblock In \emph{Proceedings of the IEEE/CVF international conference on computer vision}, pages 2639--2650, 2023.

\end{thebibliography}
}


\end{document}